\documentclass[lettersize,journal]{IEEEtran}
\usepackage{amsmath,amsfonts}
\usepackage{algorithm}
\usepackage{array}
\usepackage[caption=false,font=normalsize,labelfont=sf,textfont=sf]{subfig}
\usepackage{textcomp}
\usepackage{stfloats}
\usepackage{url}
\usepackage{verbatim}
\usepackage{graphicx}
\usepackage{cite}
\usepackage{orcidlink}
\usepackage{bm}

\usepackage{color, colortbl}
\definecolor{Gray}{gray}{0.9}
\usepackage{booktabs}
\usepackage{threeparttable}
\usepackage{CJKutf8}
\usepackage{multirow}
\usepackage{algorithm}
\usepackage{algpseudocode}

\def\argmin{\mathop{\rm argmin}}

\begin{document}
\begin{CJK}{UTF8}{gbsn}

\title{Black-box Prompt Tuning with Subspace Learning}

\author{Yuanhang Zheng\orcidlink{0000-0001-6224-8739}, Zhixing Tan\orcidlink{0000-0002-2426-6220}, Peng Li\orcidlink{0000-0003-1374-5979} and Yang Liu\orcidlink{0000-0002-3087-242X},~\IEEEmembership{Senior Member,~IEEE}
\thanks{This work is supported by the National Key R\&D Program of China (2022ZD0160502), the National Natural Science Foundation of China (No. 61925601 and 62006138), and the National Social Science Fund of China (20\&ZD279). \emph{(Corresponding authors: Zhixing Tan and Yang Liu.)}}
\thanks{Yuanhang Zheng is with Department of Computer Science and Technology, Tsinghua University, Beijing, China (e-mail: zheng-yh19@mails.tsinghua.edu.cn).}
\thanks{Zhixing Tan is with Zhongguancun Laboratory, Beijing, P.R.China (email: zxtan@zgclab.edu.cn).}
\thanks{Peng Li is with Institute for AI Industry Research (AIR), Tsinghua University, Beijing, China and Shanghai Artificial Intelligence Laboratory, Shanghai, China (e-mail: lipeng@air.tsinghua.edu.cn).}
\thanks{Yang Liu is with Department of Computer Science and Technology, Tsinghua University, Beijing, China, Institute for AI Industry Research (AIR), Tsinghua University, Beijing, China, Shanghai Artificial Intelligence Laboratory, Shanghai, China and Jiangsu Collaborative Innovation Center for Language Competence, Jiangsu, China (e-mail: liuyang2011@tsinghua.edu.cn).}
\thanks{\copyright 2024 IEEE.  Personal use of this material is permitted.  Permission from IEEE must be obtained for all other uses, in any current or future media, including reprinting/republishing this material for advertising or promotional purposes, creating new collective works, for resale or redistribution to servers or lists, or reuse of any copyrighted component of this work in other works.}
}

\markboth{IEEE/ACM Transactions on Audio, Speech, and Language Processing}
{Zheng \MakeLowercase{\textit{et al.}}: Black-box Prompt Tuning with Subspace Learning}

\maketitle

\begin{abstract}
Black-box prompt tuning employs derivative-free optimization algorithms to learn prompts within low-dimensional subspaces rather than back-propagating through the network of Large Language Models (LLMs). Recent studies reveal that black-box prompt tuning lacks versatility across tasks and LLMs, which we believe is related to the suboptimal choice of subspaces. In this paper, we introduce \underline{B}lack-box prompt tuning with \underline{S}ubspace \underline{L}earning (BSL) to enhance the versatility of black-box prompt tuning. Based on the assumption that nearly optimal prompts for similar tasks reside in a common subspace, we propose identifying such subspaces through meta-learning on a collection of similar source tasks. Consequently, for a target task that shares similarities with the source tasks, we expect that optimizing within the identified subspace can yield a prompt that performs well on the target task. Experimental results confirm that our BSL framework consistently achieves competitive performance across various downstream tasks and LLMs.
\end{abstract}


\section{Introduction}

\IEEEPARstart{I}{n} recent years, Large Language Models (LLMs) have achieved tremendous success and demonstrated their potential to generalize well across various downstream tasks~\cite{brown2020language,raffel2019exploring}. With prompt-based tuning methods~\cite{li2021prefixtuning,lester2021power}, LLMs can be easily adapted to downstream tasks by introducing only a small portion of tunable parameters. These methods also enable mix-task inference and achieve competitive performance~\cite{li2021prefixtuning,lester2021power}, making them a suitable choice for users to customize LLMs in the emerging LLM service scenario. For example, the recently launched Amazon Bedrock provides a prompt-learning API that allows users to customize the LLM service.\footnote{\url{https://aws.amazon.com/bedrock}}
 
Recently, a new form of prompt tuning methods, known as \emph{black-box prompt tuning}, has gained increasing attention. A notable characteristic of black-box prompt tuning methods is their ability to learn prompts without relying on derivative information. This is achieved through the use of derivative-free optimization (DFO) algorithms, which are also referred to as black-box optimization algorithms~\cite{sun2022blackbox,sun2022bbtv2}. Therefore, black-box prompt tuning methods eliminate the need for the backward pass and the storage of intermediate activations during the forward pass of LLMs\cite{chen2018neural}, making them efficient in computation and memory. The primary challenge with black-box tuning is the slow convergence rate of DFO algorithms in high-dimensional prompt search spaces \cite{conn2009introduction}. To mitigate this, black-box prompt tuning methods typically optimize prompts within a low-dimensional subspace of the full search space \cite{sun2022bbtv2, sun2022blackbox}.  Sun \emph{et al.} \cite{sun2022bbtv2, sun2022blackbox} demonstrate that optimizing prompts in a randomly selected affine subspace with DFO algorithms can achieve competitive performance on classification tasks under few-shot settings when compared with derivative-based prompt tuning methods. Unfortunately, despite their success, black-box prompt tuning methods have shown poor versatility across tasks and LLMs \cite{sun2022bbtv2}.

We hypothesize that the lack of versatility of black-box prompt tuning is related to the suboptimal choice of low-dimensional affine subspaces. Recent studies~\cite{sun2022bbtv2,sun2022blackbox} have found that different ways to choose affine subspaces can significantly affect the performance of LLMs on the same downstream task. Intuitively, if a subspace contains a nearly optimal prompt, optimizing in the subspace should achieve comparable performance to optimizing in the full search space. However, identifying such ``satisfying'' subspaces is difficult. First, the distribution of satisfying subspaces is generally model-specific. Sun \emph{et al.}~\cite{sun2022bbtv2} observe that choosing a random subspace according to a uniform distribution can only work well on the RoBERTa model, which implies that the distribution of satisfying subspaces differs for different LLMs. Second, the choice of subspaces is task-specific. Learning a unified subspace and tuning a low-dimensional prompt cannot achieve comparable performance with prompt tuning methods on downstream tasks \cite{qin2021exploring}.

\begin{figure*}[t]
    \centering
    \includegraphics[width=0.9\textwidth]{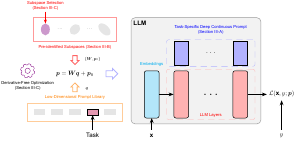}
    \caption{In our BSL framework, deep continuous prompts are optimized within a selected low-dimensional subspace via reparameterization and are learned using derivative-free optimization (DFO) algorithms. We begin by identifying common low-dimensional subspaces that contain satisfactory solutions for source tasks through meta-learning. Subsequently, the identified subspace is utilized for black-box prompt tuning on target tasks that exhibit similarities to the source tasks.}
    \label{fig:arch}
\end{figure*}

In this paper, we propose Black-box prompt tuning with Subspace Learning (BSL), a framework that aims to make black-box prompt tuning versatile across tasks and LLMs. To achieve competitive performance, BSL leverages deep continuous prompts \cite{li2021prefixtuning} to steer LLMs towards downstream tasks. The key concept behind BSL is subspace learning, which is based on the assumption that there exist nearly optimal prompts for similar tasks within a shared subspace. This assumption is supported by recent studies on prompt transferability \cite{vu2022spot, su2022transferability}. To identify such subspaces, we introduce a meta-learning algorithm that trains subspace parameters using a set of similar source tasks. As a result, the identified subspaces are both task-specific and model-specific. Consequently, for a target task that exhibits similarities with the source tasks, we expect that optimizing within the subspace will yield a prompt that performs well on the target task. Figure~\ref{fig:arch} provides an overview of our BSL framework. Since the customization of LLMs using BSL only requires forward computation, BSL is particularly suitable as an API for LLM services. Service providers can pre-identify subspaces, which can later be reused by many users when customizing the LLM service.

We evaluate our method on various types of tasks, including text classification, machine translation quality estimation, and text generation tasks. The experimental results demonstrate that our method significantly outperforms previous black-box prompt tuning methods. Moreover, it achieves competitive results when compared with derivative-based prompt tuning methods, regardless of the tasks and backbone LLMs.

\section{Preliminaries}

\subsection{Prompt Tuning}

Prompting involves adding extra tokens to the input sequence of an LLM to guide the model toward performing specific downstream tasks~\cite{brown2020language}. Depending on the representation of prompts, prompting can be categorized into \emph{textual prompting methods}~\cite{brown2020language,schick2021just,shin2020autoprompt,gao2021making} and \emph{continuous prompting methods}~\cite{li2021prefixtuning,lester2021power}.

Prompt tuning is a type of continuous prompting method. Unlike textual prompting methods, which use discrete textual tokens, prompt tuning employs continuous vectors, also known as “continuous prompts,” to steer LLMs toward downstream tasks. These learnable continuous vectors offer more flexibility than fixed token embeddings. Recent studies~\cite{li2021prefixtuning,lester2021power} also show that prompt tuning can narrow the performance gap between prompting and full fine-tuning.

\subsection{Black-Box Prompt Tuning}

Black-box prompt tuning aims to learn continuous prompts without directly computing derivative~\cite{sun2022bbtv2,sun2022blackbox}. To efficiently train prompts using DFO algorithms, the search for optimal prompts is restricted to a low-dimensional subspace of the full search space. Formally, given a continuous prompt $\bm{p}$ with dimensionality $D = N \times h$, where $h$ is the hidden size of the model and $N$ is the length of the prompt, and assuming $d < D$, the vector $\bm{p}$ can be reparameterized as:

\begin{equation}
\label{equ:prompt}
    \bm{p}= \bm{W}\bm{q}+\bm{p}_0,
\end{equation}
where $\bm{W} \in \mathbb{R}^{D \times d}$ is a \emph{projection matrix}, $\bm{p}_0 \in \mathbb{R}^{D}$ is an \emph{initial prompt}, and $\bm{q} \in \mathbb{R}^{d}$ is a trainable \emph{low-dimensional prompt}. Given $\bm{W}$ and $\bm{p}_0$, Eq.~\eqref{equ:prompt} characterizes a $d$-dimensional affine subspace within the full $\mathbb{R}^{D}$ search space.
The selection of the projection matrix varies among previous studies. Sun \emph{et al.}~\cite{sun2022blackbox} randomly generate $\bm{W}$ from uniform distributions, while Sun \emph{et al.}~\cite{sun2022bbtv2} sample $\bm{W}$ from normal distributions.

Once the low-dimensional affine subspace is chosen, the trainable prompt $\bm{q}$ can be optimized using DFO algorithms such as the Covariance Matrix Adaptation Evolution Strategy (CMA-ES)~\cite{hansen2001completely,hansen2003reducing} or the Natural Evolution Strategy (NES)~\cite{wierstra2014natural}.

\section{Approach}

In this section, we provide a detailed description of our proposed BSL approach. We begin by introducing the deep continuous prompts we employ to customize LLMs in Section~\ref{sec:prompting}. Next, we outline our proposed algorithm for learning the subspace in Section~\ref{sec:maml}. Finally, we describe the black-box optimization process of BSL in Section~\ref{sec:blackbox}.

\subsection{Deep Continuous Prompts}\label{sec:prompting}

We adopt deep continuous prompts~\cite{li2021prefixtuning} for steering LLMs to downstream tasks,\footnote{For classification tasks and regression tasks, we also use a language modeling head following Liu \emph{et al.}~\cite{liu2021ptuning}.} which introduce continuous vectors that are prepended to the hidden activations across all layers of a given LLM. For an $L$-layered LLM with a hidden size $h$, the dimension of a deep continuous prompt is $D=NLh$.

Due to the high dimensionality of deep continuous prompts, we optimize them within low-dimensional subspaces via reparameterization, as formulated in Eq.~\eqref{equ:prompt}. Unlike the approach taken by Sun \emph{et al.}~\cite{sun2022bbtv2}, we optimize prompts in $\mathbb{R}^{d}$ rather than $\mathbb{R}^{Ld}$, significantly reducing the number of tunable parameters required.

\begin{figure*}[t]
    \centering
    \includegraphics[width=0.9\textwidth]{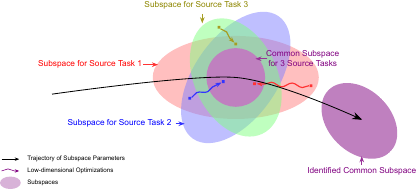}
    \caption{The trajectory of the subspace parameters during training is represented by the black line, with each point indicating a different subspace. We use meta-learning to identify a common low-dimensional subspace that contains prompts effective for the three source tasks.}
    \label{fig:metasubspace}
\end{figure*}

\subsection{Subspace Learning}\label{sec:maml}

Intuitively, a ``satisfying'' subspace should contain prompts that can effectively address a given downstream task. Since we are unaware of the specific downstream task during the subspace learning stage, the learned subspace should be versatile enough to accommodate a wide range of novel tasks. Fortunately, recent studies~\cite{vu2022spot,su2022transferability} show that prompts learned on a source task can be transferred with success to target tasks that are similar to the source task. This suggests that it is possible to identify a common subspace that is applicable to a group of similar tasks.

Based on the above intuition, we propose a meta-learning algorithm for identifying a common subspace given a set of similar tasks. Let $\mathcal{S}$ denote a set of source tasks, our goal is to learn a common subspace that contains satisfactory solutions for the source tasks. At each iteration, we first sample $m$ tasks $\{\mathcal{T}_1, \ldots, \mathcal{T}_m\}$ from $\mathcal{S}$. Then, for each task $\mathcal{T}_i$, we randomly sample a dataset $\mathcal{D}_i$ and optimize the prompts on $\mathcal{D}_i$ in the previously learned subspace. After learning prompts for all $m$ tasks, we adjust the parameters of the subspace to cover the prompts learned for different tasks. Figure~\ref{fig:metasubspace} provides an illustration.

\begin{algorithm}[t]
\caption{Meta-Learning for Subspaces Identification\label{alg:ld_maml}}
{\bf Input:} a set of source tasks $\mathcal{S}$ \\
{\bf Output:} a projection matrix $\bm{W}$ and an initial prompt $\bm{p}_0$
\begin{algorithmic}[1]
\State Randomly initialize $\bm{W}$ and $\bm{p}_0$
\While {not converged}
    \State Sample $m$ tasks $\{\mathcal{T}_1,\ldots,\mathcal{T}_{m}\}$ from $\mathcal{S}$
    \For {$i \leftarrow 1$ to $m$}
        \State Sample a dataset $\mathcal{D}_i$ from $\mathcal{T}_i$
        \State Set the low-dimensional prompt $\bm q$ to $\mathbf{0}$
        \State $\bm q_i \leftarrow \bm q - \alpha \nabla_{\bm q} \mathcal{L} (\mathcal{D}_{i};\bm{W}\bm{q}+\bm{p}_0)$
        \State $\bm{p}_i \leftarrow \bm{W} \bm{q}_i+\bm{p}_0$
        \State Sample a dataset $\mathcal{D}_i'$ from $\mathcal{T}_i$ for learning $\bm{p}_0$ and $\bm{W}$
    \EndFor
    \State $\bm{W} \leftarrow \bm{W} - \frac{\beta}{m} \sum_{i=1}^{m} \nabla_{\bm{W}} \mathcal{L}(\mathcal{D}_{i}';\bm{W} \bm{q}_i+\bm{p}_0)$ 
    \State $\bm{p}_0 \leftarrow \bm{p}_0 - \frac{\beta}{m} \sum_{i=1}^{m} \nabla_{\bm{p}_i} \mathcal{L}(\mathcal{D}_{i}';\bm{p}_i)$
\EndWhile
\State \Return $\bm{W}$ and $\bm{p}_0$ 
\end{algorithmic}
\end{algorithm}

Formally, we initially set $\bm{q}$ to $\mathbf{0}$ and then update $\bm{q}$ using gradient descent according to the following equation:
\begin{equation}
    \label{equ:phi}
    \bm{q}_i = \bm{q} - \alpha \nabla_{\bm{q}} \mathcal{L}(\mathcal{D}_i; \bm{W}\bm{q} + \bm{p}_0),
\end{equation}
where $\alpha$ denotes the learning rate for updating $\bm{q}$. Since the parameters of the backbone LLM are fixed and only prompts are tunable, we use $\mathcal{L}(\mathcal{D}; \bm{p})$ to denote the loss on the dataset $\mathcal{D}$ with prompt $\bm{p}$. We use $\bm{q}_i$ and maintain another dataset $\mathcal{D}_i'$ for learning the parameters of the subspace thereafter. Formally, the updating equations are described as
\begin{align}
  \bm{W} &\leftarrow \bm{W} - \frac{\beta}{m} \sum_{i=1}^{m} \nabla_{\bm{W}} \mathcal{L}(\mathcal{D}_i'; \bm{W}\bm{q}_i + \bm{p}_0), \label{equ:second_order_2} \\
  \bm{p}_0 &\leftarrow \bm{p}_0 - \frac{\beta}{m} \sum_{i=1}^{m} \nabla_{\bm{p}_0} \mathcal{L}(\mathcal{D}_i'; \bm{W}\bm{q}_i + \bm{p}_0), \label{equ:second_order_1}
\end{align}
where $\beta$ is the learning rate for updating $\bm{W}$ and $\bm{p}_0$. However, Eq.~\eqref{equ:second_order_1} requires the calculation of a second-order derivative with respect to $\bm{p}_0$, which significantly increases the computational overhead.\footnote{For more details, please refer to Appendix~\ref{sec:second_order}.} To alleviate this issue, we follow Finn \emph{et al.}~\cite{finn2017modelagnostic} and omit the second-order term, using the first-order approximation. Formally, let $\bm{p}_i$ denote $\bm{W}\bm{q}_i + \bm{p}_0$, and we optimize $\bm{p}_0$ using the following equation:
\begin{equation}
     \bm{p}_0 \leftarrow \bm{p}_0 - \frac{\beta}{m} \sum_{i=1}^{m} \nabla_{\bm{p}_i} \mathcal{L}(\mathcal{D}_i'; \bm{p}_i).
\end{equation}
Finally, we arrive at an identified subspace when the above procedure converges. The detailed procedure is given in Algorithm~\ref{alg:ld_maml}.

\subsection{Black-Box Optimization}\label{sec:blackbox}

Once we have identified a set of satisfactory subspaces, we can utilize them by optimizing prompts within a selected subspace using derivative-free optimization. First, for a novel target task, we need to select an appropriate subspace from the set of pre-identified subspaces. Then, we employ DFO algorithms to learn prompts within the selected subspace.

\subsubsection{Subspace Selection}\label{sec:transfer}

Given the set of pre-identified subspaces and a novel target task $\mathcal{T}$, the first step in the black-box optimization stage is to select a subspace that is suitable for the target task. We can consider two criteria for subspace selection: by task type and by inference performance.

\paragraph{Task Type}
\label{sec:subspae_task_type}

Previous studies~\cite{vu2022spot,su2022transferability} indicate that prompts are more sensitive to task types, and prompts within the same task type can transfer well to one another. Therefore, we can select a subspace that is trained on source tasks of the same task type as the target task.

\paragraph{Inference Performance}
\label{sec:subspae_inference}

Sometimes it may be challenging to select a subspace based solely on the task type. In such cases, we can also use the inference performance on a small separate development set as an indicator for subspace selection. Specifically, we evaluate the performance of the initial prompt $\bm{p}_0$ on the development set for each pre-identified candidate subspace and select the subspace that yields the best performance.

\subsubsection{Optimization with DFO Algorithms}

After the subspace is chosen, the next step is to find a low-dimensional prompt $\bm{q}^{*}$ that minimizes a given loss function $\mathcal{L}$ on the target task $\mathcal{T}$. Formally, this can be described as
\begin{equation}
\bm{q}^{*} = \argmin_{\bm{q}}\mathcal{L}(\mathcal{D};\bm{W}\bm{q}+\bm{p}_0),
\end{equation}
where $\mathcal{D}$ is the training set of the target task $\mathcal{T}$.

Since $\bm{q}$ is the only tunable parameter and is a low-dimensional vector, it can be efficiently optimized using DFO algorithms, such as the Covariance Matrix Adaptation Evolution Strategy (CMA-ES)~\cite{hansen2001completely,hansen2003reducing} algorithm or the Natural Evolution Strategy (NES)~\cite{wierstra2014natural} algorithm. Once the DFO algorithm converges, we compute the deep continuous prompt $\bm{p}^{*}$ using the equation
\begin{equation}
\bm{p}^{*} = \bm{W}\bm{q}^{*} + \bm{p}_0,
\end{equation}
to steer the LLM to the target task.

\begin{table}[t]
\centering
\caption{Dataset Statistics for Source Tasks}\label{tab:source_tasks}
\begin{tabular}{llr}
\toprule
\bf{Task} & \bf{Dataset} & \bf{Size} \\\midrule
\multirow{4}{*}{Classification} & SST-2 & 25.0k \\
& TweetEval & 24.9k \\
& IMDB & 24.8k \\
& Amazon & 25.0k \\\midrule
\multirow{9}{*}{QE} & En-Cs & 20.5k \\
& En-Fi & 6.7k \\
& En-Ru & 17.4k \\
& En-Tr & 2.0k \\
& Cs-En & 11.6k \\
& De-En & 21.7k \\
& Fi-En & 15.2k \\
& Tr-En & 17.3k \\
& Zh-En & 26.4k \\\midrule
\multirow{8}{*}{Generation} & airport & 2.8k \\
& astronaut & 1.5k \\
& building & 2.4k \\
& city & 3.0k \\
& comics character & 749 \\
& food & 3.6k \\
& monument & 786 \\
& sports team & 2.0k \\
& written work & 2.5k \\
\bottomrule
\end{tabular}
\end{table}

\section{Experiments}
\label{sec:experiments}

\subsection{Setup}
\label{sec:setup}

\subsubsection{Datasets}\label{sec:datasets} We conduct experiments on text classification, machine translation quality estimation, and text generation tasks. Since the amount of training data is usually small in LLM service scenarios~\cite{sun2022bbtv2,sun2022blackbox}, we follow Sun \emph{et al.}~\cite{sun2022blackbox} and use few-shot settings for all target tasks. The number of training and development samples is set to 128.

For text classification tasks, we conduct our experiments on sentiment classification tasks across different domains. The source tasks are constructed using the data originated from SST-2~\cite{socher2013recursive}, TweetEval~\cite{rosenthal2017semeval}, IMDB~\cite{maas2011learning}, and Amazon~\cite{zhang2015characterlevel}. For SST-2 and Amazon, we randomly select 25,000 training samples from each dataset to balance the sizes of the different datasets within the source tasks. For TweetEval, we remove all training samples labeled as ``neutral'' to unify the labels across all source tasks. During the execution of our proposed meta-learning algorithm, to assess the performance of the subspace on unseen tasks, we assess the learned subspace on the MPQA~\cite{wiebe2005annotating} dataset.\footnote{During the assessment process, we fine-tune the low-dimensional prompt $\bm{q}$ on a few-shot training dataset with 128 samples while keeping $\bm{W}$ and $\bm{p}_0$ fixed, and then assess the performance on the test dataset.} We use the data originated from Yelp~\cite{zhang2015characterlevel}, CR~\cite{hu2004mining}, and MR~\cite{pang2005seeing} to construct the target tasks. 

For machine translation quality estimation (QE) tasks, we select 9 QE tasks with different language pairs (En-Cs, En-Fi, En-Ru, En-Tr, Cs-En, De-En, Fi-En, Tr-En, and Zh-En) from the WMT 2017 ``QE as a metric'' as source tasks. During the meta-learning process, we use the data of En-De and Ru-En from the WMT 2020 QE dataset for assessing the performance of the subspace. We use the WMT 2020 English-Chinese (En-Zh), Nepali-English (Ne-En), and Romanian-English (Ro-En) QE as target tasks. 

For the text generation task, we construct the source tasks using the WebNLG~\cite{gardent2017webnlg} dataset, which covers multiple domains. Specifically, we learn the subspace on data from 9 different domains (airport, astronaut, building, city, comics character, food, monument, sports team, and written work) and assess the performance of the subspace on the university domain during meta-learning. We use the E2E~\cite{novikova2017e2e} dataset to construct the target task.

The statistics of the datasets are presented in Table~\ref{tab:source_tasks}, Table~\ref{tab:evaluation_tasks}, and Table~\ref{tab:target_tasks}.

\subsubsection{Baselines} We compare our BSL approach with both derivative-based tuning methods and black-box prompt tuning methods:
\begin{itemize}
  \item PLM: The pre-trained language model is directly used to infer on the downstream tasks without using any fine-tuning or prompt-based tuning methods.
  \item Fine-Tuning: A method that fine-tunes all the parameters of the LLM on the downstream tasks using gradient descent.
  \item Prefix-Tuning~\cite{li2021prefixtuning}: A prompt-based tuning method that utilizes deep continuous prompts to guide LLMs towards downstream tasks. The prompts are also learned through gradient descent.
  \item BBT~\cite{sun2022blackbox}: A black-box prompt tuning method that optimizes prompts within a low-dimensional subspace, with the projection matrix of the subspace sampled from a uniform distribution.
  \item BBTv2~\cite{sun2022bbtv2}: A black-box prompt tuning method that uses deep continuous prompts, optimizing the sub-prompt of each layer independently within different subspaces. The projection matrices for these subspaces are sampled from a normal distribution.
\end{itemize}

\begin{table}[t]
\centering
\caption{Dataset Statistics for Tasks Used in Assessment}\label{tab:evaluation_tasks}
\begin{tabular}{llrr}
\toprule
\multirow{2}{*}{\bf{Task}} & \multirow{2}{*}{\bf{Dataset}} & \multicolumn{2}{c}{\bf{Size}} \\\cmidrule{3-4}
& & Train & Test \\\midrule
Classification & MPQA & 128 & 1.8k \\\midrule
\multirow{2}{*}{QE} & En-De & 128 & 1.0k \\
& Ru-En & 128 & 1.0k \\\midrule
Generation & university & 128 & 90 \\
\bottomrule
\end{tabular}
\end{table}

For fair comparisons, we re-implement the baseline methods. For all the prompt-based baseline methods, we \emph{pre-train an initial prompt on the same source tasks as used in BSL}. Subsequently, we conduct the  optimization process (either derivative-based optimization or derivative-free optimization) on the training set of the target tasks. In our re-implementation, we utilize deep continuous prompts for both BBT and BBTv2. We employ two distinct pre-training strategies for the prompt-based baseline methods: (1) pre-training the prompt on a mixed dataset that encompasses the datasets of all source tasks and (2) pre-training separate source prompts individually on each source task, then selecting the best source prompt based on the ``inference performance'' criteria proposed in Section~\ref{sec:subspae_inference}. We denote these pre-training strategies as ``mixed'' and ``multi'', respectively. For the baseline method Fine-Tuning, we also utilize the ``mixed'' and ``multi'' pre-training strategies. However, we adjust the parameters of the entire model in the case of Fine-Tuning.

\begin{table}[t]
\centering
\caption{Dataset Statistics for Target Tasks}\label{tab:target_tasks}
\setlength{\tabcolsep}{5pt}
\begin{tabular}{llrrr}
\toprule
\multirow{2}{*}{\bf{Task}} & \multirow{2}{*}{\bf{Dataset}} & \multicolumn{3}{c}{\bf{Size}} \\\cmidrule{3-5}
& & Train & Dev & Test \\\midrule
\multirow{3}{*}{Classification} & Yelp & 128 & 128 & 38.0k \\
& CR & 128 & 128 & 2.0k \\
& MR & 128 & 128 & 2.0k \\\midrule
\multirow{3}{*}{QE} & En-Zh & 128 & 128 & 1.0k \\
& Ne-En & 128 & 128 & 1.0k \\
& Ro-En & 128 & 128 & 1.0k \\\midrule
Generation & E2E & 128 & 128 & 630 \\
\bottomrule
\end{tabular}
\end{table}

\subsubsection{Evaluation Metrics} We employ distinct evaluation metrics for each task. Specifically, we use accuracy for text classification tasks. For QE tasks, we utilize Pearson's correlation coefficient as the evaluation metric. For the text generation task, we report BLEU~\cite{papineni2002bleu}, METEOR~\cite{banerjee2005meteor}, and ROUGE-L~\cite{lin2004rouge} to evaluate the quality of the generated text.

\subsubsection{Implementation Details} Since our experiments are conducted on the few-shot settings, we conduct all experiments (except the experiments of the baseline method PLM) using three different random seeds and present the average performance and standard deviation for all tasks. For the experiments of the baseline method PLM, we report the results of a single run, since the pre-trained language model is directly used for inference without any fine-tuning or prompt tuning in the baseline method PLM. We implement all methods on top of the Transformers library~\cite{wolf2020transformers}. For the backbone LLMs, we use \texttt{bert-base-uncased}~\cite{devlin2019bert} for text classification tasks, multilingual BERT~\cite{devlin2019bert} for QE tasks, and GPT-2$_{\textrm{medium}}$~\cite{radford2019language} for the generation task. We set the prompt length $N$ to 30 and the dimensionality $d$ of the low-dimensional prompt $\bm{q}$ to 500.

During the meta-learning process, we utilize the Adam~\cite{kingma2015adam} optimizer with parameters $\beta_1=0.9$, $\beta_2=0.999$, and $\epsilon=10^{-8}$. The learning rate is set to $\alpha=3 \times 10^{-4}$ for both $\bm{q}$ and $\bm{W}$, and $\beta=3 \times 10^{-4}$ for $\bm{p}_0$. For each iteration, we sample two source tasks (i.e., $m$ is set to 2). The size of the sampled datasets $D_i$ and $D_i'$ is set to 64. When updating the low-dimensional prompt $\bm{q}$ in the meta-learning process (Eq.\eqref{equ:phi}), we follow Finn \emph{et al.}~\cite{finn2017modelagnostic} and optimize $\bm{q}$ for multiple steps. Specifically, we perform 16 gradient steps for updating $\bm{q}$ and set the batch size to 8 (for text classification and QE tasks) or 4 (for generation tasks). We assess the performance of the subspace every 200 iterations and select the subspace with the best performance.

Before the black-box optimization stage, we select the trained subspace based on the ``task type'' criterion introduced in Section~\ref{sec:subspae_task_type}, as the task types used in our experiments (i.e., classification, QE, and generation) are significantly distinct. During the black-box optimization stage, we adhere to the approach of Sun \emph{et al.}~\cite{sun2022bbtv2,sun2022blackbox} and employ CMA-ES~\cite{hansen2001completely,hansen2003reducing} as the DFO algorithm. We restrict the number of API calls (i.e., the number of forward computations) to 8,000. We set the population size of the CMA-ES algorithm to 20, which results in a maximum of 400 optimization steps. We evaluate the performance on the development set every 20 optimization steps.

\subsection{Main Results}

\begin{table}[t]
    \centering
    \caption{Results on Text Classification Tasks}\label{tab:main}
    \begin{threeparttable}
    \begin{tabular}{lcccc}
        \toprule
        \textbf{Method} & \textbf{Params.} & \textbf{Yelp} & \textbf{CR} & \textbf{MR} \\\midrule
        PLM & 0 & 52.3 & 49.9 & 50.1 \\
        Fine-Tuning (mixed) & 134M & 92.3\tiny{$\pm$0.8} & 88.5\tiny{$\pm$1.2} & 86.1\tiny{$\pm$2.0} \\
        Fine-Tuning (multi) & 134M & 91.5\tiny{$\pm$1.3} & 87.7\tiny{$\pm$1.0} & 83.5\tiny{$\pm$0.5} \\\midrule
        \multicolumn{5}{c}{\emph{derivative-based prompt tuning methods}} \\\midrule
        Prefix-Tuning (mixed) & 276k & 90.5\tiny{$\pm$0.2} & 88.5\tiny{$\pm$0.3} & 85.3\tiny{$\pm$0.2} \\
        Prefix-Tuning (multi) & 276k & 92.9\tiny{$\pm$0.6} & 87.6\tiny{$\pm$0.4} & 81.5\tiny{$\pm$1.5} \\\midrule
        \multicolumn{5}{c}{\emph{black-box prompt tuning methods}} \\\midrule
        BBT (mixed) & 500 & 93.7\tiny{$\pm$0.1} & 88.6\tiny{$\pm$0.6} & 84.6\tiny{$\pm$0.1} \\
        BBT (multi) & 500 & 93.8\tiny{$\pm$0.2} & 87.8\tiny{$\pm$0.4} & 85.2\tiny{$\pm$0.3} \\
        BBTv2 (mixed) & 6k & 93.8\tiny{$\pm$0.1} & 88.4\tiny{$\pm$0.4} & 84.7\tiny{$\pm$0.2} \\
        BBTv2 (multi) & 6k & 93.9\tiny{$\pm$0.1} & 88.0\tiny{$\pm$0.4} & 85.5\tiny{$\pm$0.6} \\
        BSL (\emph{Ours}) & 500 & \textbf{94.6}\tiny{$\pm$0.1} & \textbf{88.7}\tiny{$\pm$0.6} & \textbf{88.0}\tiny{$\pm$0.2} \\
        \bottomrule
    \end{tabular}
    \smallskip
    \footnotesize ``Params.'' denotes the number of trainable parameters during the optimization process.
    \end{threeparttable}
\end{table}

\begin{table}[t]
    \centering
    \caption{Results on Machine Translation Quality Estimation Tasks}\label{tab:main_qe}
    \begin{threeparttable}
    \begin{tabular}{lcccc}
        \toprule
        \textbf{Method} & \textbf{Params.} & \textbf{En-Zh} & \textbf{Ne-En} & \textbf{Ro-En} \\\midrule
        PLM & 0 & 4.6 & 35.3 & 43.6 \\
        Fine-Tuning (mixed) & 270M & 42.7\tiny{$\pm$2.4} & 63.7\tiny{$\pm$2.6} & 81.5\tiny{$\pm$1.0} \\
        Fine-Tuning (multi) & 270M & 41.1\tiny{$\pm$3.1} & 57.9\tiny{$\pm$4.0} & 78.1\tiny{$\pm$0.4} \\\midrule
        \multicolumn{4}{c}{\emph{derivative-based prompt tuning methods}} \\\midrule
        Prefix-Tuning (mixed) & 276k & 35.9\tiny{$\pm$1.3} & 54.3\tiny{$\pm$1.2} & 76.1\tiny{$\pm$1.9} \\
        Prefix-Tuning (multi) & 276k & 28.7\tiny{$\pm$6.9} & 51.3\tiny{$\pm$1.1} & 72.8\tiny{$\pm$2.2} \\\midrule
        \multicolumn{4}{c}{\emph{black-box prompt tuning methods}} \\\midrule
        BBT (mixed) & 500 & 34.3\tiny{$\pm$0.6} & 54.7\tiny{$\pm$1.4} & 74.0\tiny{$\pm$0.5} \\
        BBT (multi) & 500 & 29.5\tiny{$\pm$1.1} & 52.6\tiny{$\pm$1.9} & 70.2\tiny{$\pm$2.4} \\
        BBTv2 (mixed) & 6k & 35.4\tiny{$\pm$1.6} & 54.3\tiny{$\pm$1.5} & 74.1\tiny{$\pm$2.0} \\
        BBTv2 (multi) & 6k & 31.1\tiny{$\pm$3.1} & 54.3\tiny{$\pm$3.0} & 71.2\tiny{$\pm$1.4} \\
        BSL (\emph{Ours}) & 500 & \textbf{36.3}\tiny{$\pm$0.4} & \textbf{57.4}\tiny{$\pm$4.7} & \textbf{77.6}\tiny{$\pm$1.1} \\
        \bottomrule
    \end{tabular}
    \smallskip
    \footnotesize ``Params.'' denotes the number of trainable parameters during the optimization process.
    \end{threeparttable}
\end{table}

\begin{table}[t]
    \centering
    \setlength{\tabcolsep}{5pt}
    \caption{Results on the Generation Task with Three Different Evaluation Metrics}\label{tab:main_gen}
    \begin{threeparttable}
    \begin{tabular}{lcccc}
        \toprule
        \textbf{Method} & \textbf{Params.} & \textbf{BLEU} & \textbf{METEOR} & \textbf{ROUGE-L} \\\midrule
        PLM & 0 & 0.0 & 4.2 & 17.3 \\
        Fine-Tuning (mixed) & 380M & 57.0\tiny{$\pm$3.0} & 36.5\tiny{$\pm$1.0} & 50.0\tiny{$\pm$0.6} \\
        Fine-Tuning (multi) & 380M & 55.2\tiny{$\pm$3.5} & 35.4\tiny{$\pm$1.3} & 49.4\tiny{$\pm$0.6} \\\midrule
        \multicolumn{4}{c}{\emph{derivative-based prompt tuning methods}} \\\midrule
        Prefix-Tuning (mixed) & 737k & 57.3\tiny{$\pm$4.5} & 36.4\tiny{$\pm$1.7} & 48.0\tiny{$\pm$1.2} \\
        Prefix-Tuning (multi) & 737k & 56.2\tiny{$\pm$2.6} & 35.4\tiny{$\pm$0.9} & 48.3\tiny{$\pm$0.5} \\\midrule
        \multicolumn{4}{c}{\emph{black-box prompt tuning methods}} \\\midrule
        BBT (mixed) & 500 & 38.4\tiny{$\pm$4.1} & 29.6\tiny{$\pm$1.0} & 42.4\tiny{$\pm$1.2} \\
        BBT (multi) & 500 & 38.8\tiny{$\pm$3.5} & 29.9\tiny{$\pm2.1$} & 43.5\tiny{$\pm$0.7} \\
        BBTv2 (mixed) & 12k & 40.4\tiny{$\pm$2.3} & 31.4\tiny{$\pm$0.5} & 43.4\tiny{$\pm$0.7} \\
        BBTv2 (multi) & 12k & 41.9\tiny{$\pm$0.6} & 31.7\tiny{$\pm$0.6} & 43.3\tiny{$\pm$0.5} \\
        BSL (\emph{Ours}) & 500 & \textbf{49.1}\tiny{$\pm$3.1} & \textbf{32.1}\tiny{$\pm$1.2} & \textbf{45.9}\tiny{$\pm$0.9} \\
        \bottomrule
    \end{tabular}
    \smallskip
    \footnotesize ``Params.'' denotes the number of trainable parameters during the optimization process.
    \end{threeparttable}
\end{table}

\begin{table}[t]
    \centering
    \caption{Results on different tasks before optimizing on the target tasks}\label{tab:results_before}
    \begin{threeparttable}
        \begin{tabular}{lccc}
            \toprule
            \textbf{Method} & \textbf{MR} & \textbf{Ne-En} & \textbf{E2E} \\\midrule
            Prefix-Tuning & 84.7 & 52.0 & 23.6 \\
            BSL (\emph{Ours}) & 87.0 & 54.8 & 35.7 \\
            \bottomrule
        \end{tabular}
        \smallskip
        \footnotesize We employ BLEU as the evaluation metric for the E2E task.
    \end{threeparttable}
\end{table}

\subsubsection{Results on Text Classification} Table~\ref{tab:main} shows the performance of various methods on text classification tasks. We can observe that BBTv2~\cite{sun2022bbtv2} does not consistently outperform BBT~\cite{sun2022blackbox}, despite BBTv2 having significantly more tunable parameters. However, our proposed BSL outperforms both BBT and BBTv2 on all three text classification tasks. On average, BSL outperforms BBT (mixed), BBT (multi), and BBTv2 (mixed) by 1.5 points, and BBTv2 (multi) by 1.3 points. It is noteworthy that the same subspace is utilized for all three target tasks. This suggests that our approach can learn subspaces that are well-suited to tasks with a similar task type, enabling users to select projections based on the task type.

\subsubsection{Results on QE and Generation Tasks} Tables~\ref{tab:main_qe} and~\ref{tab:main_gen} present the performance of various methods on QE and generation tasks, respectively.

For QE tasks, our proposed BSL also outperforms BBT and BBTv2 on all three target tasks. On average, BSL outperforms BBT (mixed) by 2.8 points, BBT (multi) by 6.3 points, BBTv2 (mixed) by 2.5 points, and BBTv2 (multi) by 4.9 points. We also utilize the same subspace for the three target tasks.

For the generation task, our proposed BSL outperforms the two black-box baseline methods when using three different evaluation metrics. BSL outperforms BBT (mixed) by 10.7 points, BBT (multi) by 10.3 points, BBTv2 (mixed) by 8.7 points, and BBTv2 (multi) by 7.2 points when using BLEU~\cite{papineni2002bleu} as the evaluation metric.

As we employ different backbone LLMs for different types of tasks, the results suggest that our BSL is versatile across both tasks and LLMs.

\subsubsection{Comparison between BSL and Prefix-Tuning} As observed in the experimental results presented in Tables \ref{tab:main}, \ref{tab:main_qe}, and \ref{tab:main_gen}, the performance of BSL relative to Prefix-Tuning varies depending on the task type. BSL outperforms Prefix-Tuning on text classification and QE tasks, but lags behind Prefix-Tuning on the generation task. To further investigate the performance of BSL and Prefix-Tuning, we evaluate the performance on the \emph{target tasks before applying the derivative-based or black-box optimization process}. The results are summarized in Table~\ref{tab:results_before}. According to the experimental results, the performance gap between the initial prompt and the optimized prompt is relatively narrow for classification and QE tasks but more pronounced for the generation task. This discrepancy may be attributed to the inherent complexity and variability of the output in the generation task, as compared with those in classification and QE tasks.
Consequently, BSL achieves competitive performance with the derivative-based method Prefix-Tuning on classification and QE tasks but underperforms Prefix-Tuning on the generation task. This coincides with Huang and Zhang~\cite{huang2020blackbox}, which shows that the transferability significantly affects the efficiency and the effectiveness of black-box optimization. Our proposed BSL, with its ability to learn a superior subspace for black-box prompt tuning, outperforms Prefix-Tuning, BBT, and BBTv2 on classification and QE tasks but still lags behind Prefix-Tuning on the generation task.

\subsection{Comparisons of Efficiency}

\begin{table}[t]
    \centering
    \caption{Comparison of Training Efficiency for Baselines and BSL}\label{tab:efficiency}
    \begin{tabular}{lcc}
        \toprule
        \textbf{Method} & \textbf{Training Time} & \textbf{GPU Memory} \\\midrule
        Fine-Tuning & 428s & 4392MB \\
        Prefix-Tuning & 632s & 5549MB \\
        BBT & 270s & 3233MB \\
        BBTv2 & 274s & 2917MB \\
        BSL (\emph{Ours}) & 276s & 3061MB \\
        \bottomrule
    \end{tabular}
\end{table}

\begin{figure}[t]
    \centering
    \includegraphics[trim={0.5cm 0.5cm 0.5cm 0.5cm},clip,scale=0.95]{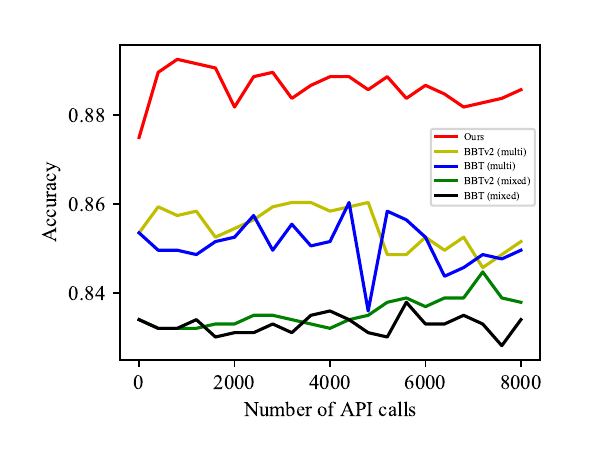}
    \caption{Learning curves of different black-box prompt tuning methods. The results are evaluated on a separate development set with 1,024 examples.}
    \label{fig:eff}
\end{figure}

\begin{figure}[t]
    \centering
    \includegraphics[trim={0.5cm 0.4cm 0.3cm 0.1cm},clip,scale=0.95]{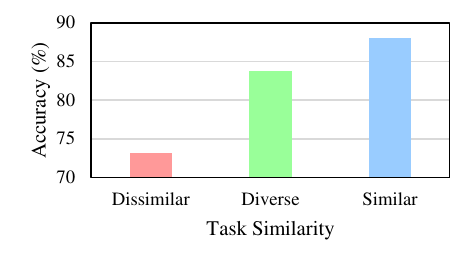}
    \caption{Effect of source tasks on BSL with regard to task similarities.}
    \label{fig:source}
\end{figure}

\begin{figure}[t]
    \centering
    \includegraphics[trim={0.5cm 0.4cm 0.3cm 0.1cm},clip,scale=0.95]{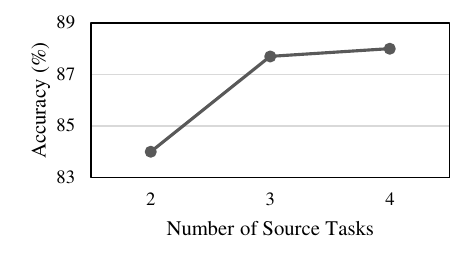}
    \caption{Effect of source tasks on BSL with regard to the number of source tasks.}
    \label{fig:source1}
\end{figure}

\subsubsection{Comparison of Training Efficiency} To compare the training efficiency of various prompt tuning methods, we conduct an experiment on a classification task built upon the MR dataset\footnote{For simplicity, we use “the MR task” to represent classification task built upon the MR dataset in the remaining part of the paper.}. During the experiment, each method runs for 8,000 steps (i.e., API calls). We measure the training time and GPU consumption for each method. Table~\ref{tab:efficiency} presents the results. We can observe that the derivative-based Prefix-Tuning method requires significantly more training time and GPU memory compared with Fine-Tuning and black-box prompt tuning methods. Although Prefix-Tuning reduces the storage required for each downstream task compared with Fine-Tuning, it still needs to back-propagate through the entire LLM network. Thus, it cannot reduce GPU memory consumption during training. Black-box baseline methods (i.e., BBT, BBTv2, and BSL), which eliminate the need for computing derivatives, are more memory-efficient. Our proposed BSL method requires a similar amount of training time and GPU memory compared with BBT and BBTv2. This indicates that our proposed BSL can enhance the performance of black-box prompt tuning without sacrificing training efficiency.

\subsubsection{Comparison of Convergence Speed} Figure~\ref{fig:eff} illustrates the learning curves of different black-box prompt tuning methods on the MR task. The performance is evaluated on a separate development set with 1,024 examples. It is evident that our BSL consistently outperforms BBT and BBTv2 across various numbers of API calls. On the one hand, our proposed BSL achieves a higher accuracy before black-box prompt tuning compared with the baselines. This aligns with Gu \emph{et al.}~\cite{gu2018metalearning}, which demonstrates that meta-learning can enhance the performance in zero-shot scenarios. On the other hand, our framework also significantly accelerates the convergence speed of the DFO algorithm. This suggests that our approach can reduce the number of required API calls, which is advantageous in the context of LLM services.

\subsection{Effect of Source Tasks}
\label{sec:auxiliary}

We examine the influence of source tasks on the learning of subspaces. We employ the MR task as the target task and create various source tasks by adjusting task similarities and the quantity of tasks.

\subsubsection{Task Similarities}\label{sec:auxiliary_task_type} We commence by examining the construction of source tasks. Based on task similarities, we consider the following variations:
\begin{itemize}
  \item Similar: All four source tasks are similar to the target task, i.e., they share the same task type as the target task. This represents the default approach for constructing source tasks in our study.
  \item Dissimilar: All four source tasks are dissimilar to the target task.
  \item Diverse: The source tasks comprise two tasks from the ``similar'' category and two tasks from the ``dissimilar'' category.
\end{itemize}

For the ``Dissimilar'' variant, we employ four distinct NLI (Natural Language Inference) tasks as source tasks to train the subspace. Specifically, we utilize SNLI~\cite{bowman2015large}, MNLI~\cite{williams2018broad}, QNLI~\cite{rajpurkar2016squad}, and RTE~\cite{dagan2005pascal,barhaim2006second,giampiccolo2007third,bentivogli2009fifth} to construct the source tasks. For SNLI and MNLI, we combine the labels ``contradictory'' and ``neutral'' into a new label ``not entailment'' to harmonize the labels across the source tasks. For the ``Diverse'' variant, we utilize two similar source tasks Amazon and TweetEval and two dissimilar source tasks MNLI and SNLI to train the subspace.

\begin{table}[t]
    \centering
    \caption{Effect of the Meta-Learning Algorithm on BSL}\label{tab:meta_learning}
    \begin{tabular}{lc}
        \toprule
        \textbf{Algorithm} & \textbf{Accuracy} \\\midrule
        ALL & 72.1\tiny{$\pm$2.6} \\
        SPC & 70.0\tiny{$\pm$4.8} \\
        INI & 74.2\tiny{$\pm$0.6} \\\midrule
        \emph{Ours} & \textbf{88.0}\tiny{$\pm$0.2} \\
        \bottomrule
    \end{tabular}
\end{table}

\begin{table}[t]
    \centering
    \caption{Effect of the DFO Algorithm on BSL}\label{tab:dfo}
    \begin{tabular}{lc}
        \toprule
        \textbf{Algorithm} & \textbf{Accuracy} \\\midrule
        CMA-ES & 88.0\tiny{$\pm$0.2} \\
        NES & 88.1\tiny{$\pm$0.2} \\
        \bottomrule
    \end{tabular}
\end{table}

As depicted in Figure~\ref{fig:source}, the performance significantly deteriorates when employing a subspace trained on dissimilar tasks. Conversely, the best performance is achieved using a subspace trained on source tasks that mirror the target task. This finding aligns with our intuition that nearly optimal prompts for similar tasks reside within a common subspace. Consequently, it is essential to select similar tasks as source tasks.

\subsubsection{Number of Source Tasks} We also delve into how the number of source tasks impacts performance on the target task. Figure~\ref{fig:source1} illustrates that performance improves with an increase in the number of source tasks. This aligns with our intuition that a greater number of source tasks aids in the learning of subspaces. With more source tasks, our proposed meta-learning algorithm can train subspaces that are more effective for target tasks.

\subsection{Analysis}

In this section, we conduct further experiments on the MR task to analyze the importance of different components in our BSL method.

\begin{table}[t]
    \centering
    \caption{Effect of Prompt Length on BSL}\label{tab:prompt_length}
    \begin{tabular}{cc}
        \toprule
        \textbf{Length} & \textbf{Accuracy} \\\midrule
        10 & 87.9\tiny{$\pm$0.4} \\
        30 & 88.0\tiny{$\pm$0.2} \\
        50 & 88.0\tiny{$\pm$0.4} \\
        \bottomrule
    \end{tabular}
\end{table}

\begin{table}[t]
    \centering
    \caption{Effect of Subspace Dimensionality on BSL}\label{tab:subspace_dimensionality}
    \begin{tabular}{cc}
        \toprule
        \textbf{Dimensionality} & \textbf{Accuracy} \\\midrule
        300 & 87.9\tiny{$\pm$0.4} \\
        500 & 88.0\tiny{$\pm$0.2} \\
        700 & 88.0\tiny{$\pm$0.5} \\
        \bottomrule
    \end{tabular}
\end{table}

\subsubsection{Meta-learning Algorithms} We first investigate the impact of meta-learning algorithms in our approach. We compare our meta-learning algorithm with various variants of the MAML algorithm~\cite{finn2017modelagnostic}. Table~\ref{tab:meta_learning} presents the results. ``ALL'' indicates that $\bm{W}$, $\bm{q}$, and $\bm{p}_0$ are jointly optimized with the MAML algorithm. ``SPC'' signifies that only the subspace parameters $\bm{W}$ and $\bm{p}_0$ are optimized with MAML, while $\bm{q}$ is fixed at a random value. ``INI'' indicates that the MAML algorithm solely learns the initial prompt $\bm{p}_0$, and both $\bm{W}$ and $\bm{q}$ are kept constant. It is evident that all three variants perform significantly worse than our proposed approach. Our meta-learning algorithm is specifically tailored for black-box prompt tuning, aiming to learn a subspace that performs well with various values of $\bm{q}$. Consequently, we demonstrate that our proposed meta-learning algorithm is crucial in learning the subspaces.

\subsubsection{DFO Algorithms} Table~\ref{tab:dfo} illustrates the results obtained using different DFO algorithms for learning the low-dimensional prompt. The experimental results indicate that the accuracy is comparable when either CMA-ES~\cite{hansen2001completely,hansen2003reducing} or NES~\cite{wierstra2014natural} is employed as the DFO algorithm. This suggests that our proposed BSL is resilient to the selection of DFO algorithms.

\subsubsection{Prompt Length and Subspace Dimensionality} As depicted in Table~\ref{tab:prompt_length} and Table~\ref{tab:subspace_dimensionality}, we also carry out experiments to investigate the impact of the prompt length and the subspace dimensionality. The experimental results reveal that the accuracy remains consistent across various prompt lengths and subspace dimensionalities, indicating that our proposed BSL is insensitive to the choice of prompt length and subspace dimensionality.

\begin{table}[t]
\centering
\caption{Accuracy on Test Sets of Target Tasks Using Subspaces Learned from Different Source Task Types}\label{tab:auto}
\begin{tabular}{lccc}
\toprule
\bf{Source Task Type} & \bf{Yelp} & \bf{CR} & \bf{MR} \\\midrule
\textbf{(a) Sentiment Classification} & \textbf{94.6}{\tiny $\pm$0.1} & \textbf{88.7}{\tiny $\pm$0.6} & \textbf{88.0}{\tiny $\pm$0.2} \\\midrule
(b) Natural Language Inference & 79.1{\tiny $\pm$2.4} & 79.0{\tiny $\pm$1.6} & 73.2{\tiny $\pm$2.2} \\
(c) Paraphrase Detection & 75.6{\tiny $\pm$4.3} & 75.2{\tiny $\pm$2.5} & 67.3{\tiny $\pm$6.7} \\
(d) Offensive Langauge Detection & 83.2{\tiny $\pm$0.4} & 78.2{\tiny $\pm$2.2} & 71.0{\tiny $\pm$2.8} \\
(e) Fact Checking & 81.8{\tiny $\pm$0.6} & 81.8{\tiny $\pm$0.7} & 68.4{\tiny $\pm$8.9} \\
\bottomrule
\end{tabular}
\end{table}

\begin{table}[t]
\centering
\caption{Accuracy on Development Sets of Target Tasks Before DFO Using Subspaces Learned from Different Source Task Types}\label{tab:auto2}
\begin{tabular}{lccc}
\toprule
\bf{Source Task Type} & \bf{Yelp} & \bf{CR} & \bf{MR} \\\midrule
\textbf{(a) Sentiment Classification} & \textbf{94.3}{\tiny $\pm$1.2} & \textbf{87.7}{\tiny $\pm$2.0} & \textbf{84.1}{\tiny $\pm$5.6} \\\midrule
(b) Natural Language Inference & 73.2{\tiny $\pm$7.9} & 61.5{\tiny $\pm$5.3} & 53.6{\tiny $\pm$4.5} \\
(c) Paraphrase Detection & 50.0{\tiny $\pm$0.0} & 50.0{\tiny $\pm$0.0} & 50.0{\tiny $\pm$0.0} \\
(d) Offensive Langauge Detection & 45.1{\tiny $\pm$1.8} & 49.7{\tiny $\pm$0.5} & 49.5{\tiny $\pm$1.8} \\
(e) Fact Checking & 50.0{\tiny $\pm$0.0} & 50.0{\tiny $\pm$0.0} & 50.0{\tiny $\pm$0.0} \\
\bottomrule
\end{tabular}
\end{table}

\begin{figure}[t]
    \centering
    \subfloat[]{
        \includegraphics[trim={0.5cm 0.4cm 0.4cm 0.3cm},clip, scale=0.9]{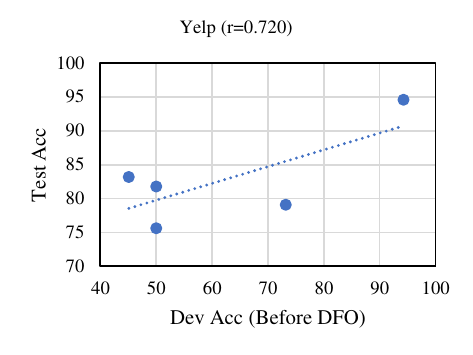}
    }
    \\
    \subfloat[]{
        \includegraphics[trim={0.5cm 0.4cm 0.4cm 0.3cm},clip, scale=0.9]{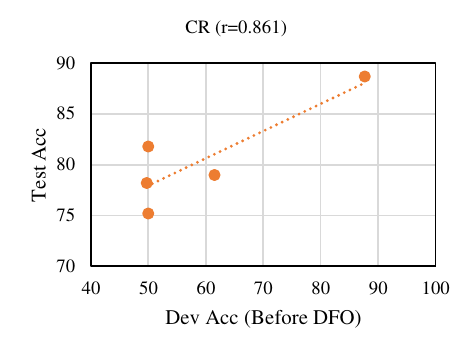}
    }
    \\
    \subfloat[]{
        \includegraphics[trim={0.5cm 0.4cm 0.4cm 0.3cm},clip, scale=0.9]{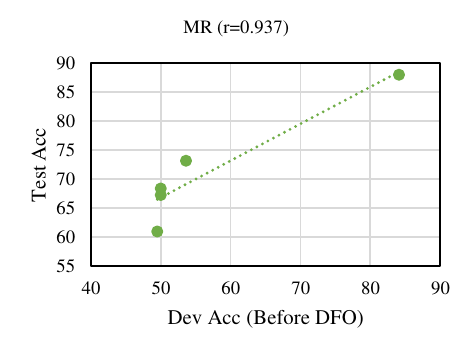}
    }
    \caption{Pearson's correlation coefficient between the accuracy on the development set before DFO and the test accuracy.}\label{fig:correl}
\end{figure}

\subsection{Experiments on Subspace Selection}
\label{sec:auto_subspace_1}

In Section~\ref{sec:transfer}, we introduce criteria for both manual and automatic subspace selection. Specifically, for manual subspace selection, we can choose the appropriate subspace based on the task type of the source and target tasks. For automatic subspace selection, we can also utilize the zero-shot performance on a separate development set as an indicator for the selection of subspaces. We can perform inference on the development set using the learned initial prompt $\bm{p}_0$ of each candidate subspace and select the subspace with the best performance on the development set.

To validate the effectiveness of the proposed criteria for subspace selection, we conduct experiments using 5 different candidate subspaces learned on various types of source tasks, including (a) sentiment analysis, (b) natural language inference, (c) paraphrase detection, (d) offensive language detection, and (e) fact checking.

Specifically, the subspace (a) corresponds to the learned subspace for sentiment classification using our proposed meta-learning algorithm, which is detailed in Section~\ref{sec:datasets}. The subspace (b) corresponds to the learned subspace for the variant Dissimilar introduced in Section~\ref{sec:auxiliary_task_type}. For the subspace (c), we utilize QQP,\footnote{\url{https://data.quora.com/First-Quora-Dataset-Release-Question-Pairs}} PAWS~\cite{paws2019naacl}, and MRPC~\cite{dolan2005automatically} to construct the source tasks. For the subspace (d), we use Wiki Toxic,\footnote{\url{https://huggingface.co/datasets/OxAISH-AL-LLM/wiki_toxic}} OLID~\cite{zampieri2019predicting}, and the Hate Speech Dataset~\cite{gibert2018hate} to construct the source tasks. For the subspace (e), we use FEVER~\cite{thorne208fever} and ADE-Corpus-V2~\cite{gurulingappa2012development} to construct the source tasks.

We first apply black-box prompt tuning to the three target text classification tasks. Table~\ref{tab:auto} demonstrates that the subspace (a) achieves the highest test accuracy. Since the target tasks are sentiment analysis tasks, we can infer that the task type significantly influences the performance of our proposed BSL and it can be considered as an effective criterion for subspace selection.

\begin{figure}[t]
    \centering
    \includegraphics[trim={0.5cm 0.4cm 0.2cm 0.1cm},clip,scale=0.7]{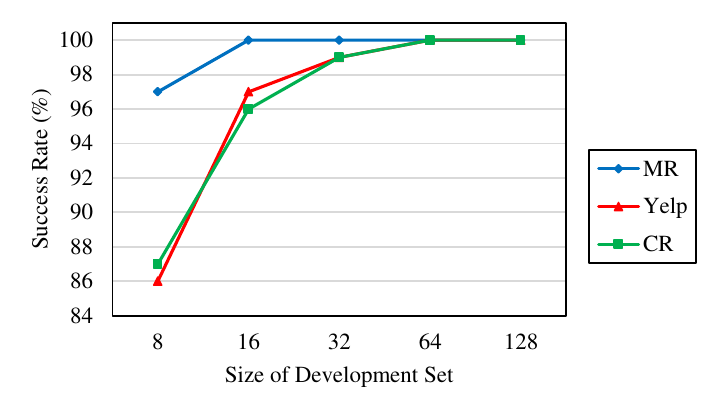}
    \caption{Success rate of identifying the optimal subspace among five candidate subspaces.}
    \label{fig:auto1}
\end{figure}

Next, we investigate the feasibility of our proposed criteria for automatic subspace selection. Specifically, for each target task, we perform inference on a development set with 128 samples using the initial prompt of each candidate subspace and then evaluate the accuracy on the development set. Table~\ref{tab:auto2} illustrates that the subspace (a), which achieves the best performance on the test set, also attains the best zero-shot performance on the development set. Moreover, as depicted in Figure~\ref{fig:correl}, we also calculate the Pearson's correlation coefficient between the zero-shot performance on the development set and the test accuracy and find that the zero-shot performance on the development set is highly correlated with the performance on the test set. This suggests that the zero-shot performance on the development set can serve as an indicator for automatic subspace selection.

Finally, we further examine the effectiveness of automatic subspace selection by attempting to identify the best subspace from the 5 candidate subspaces using development sets with varying sizes (8, 16, 32, 64, and 128 samples). We evaluate the accuracy on the development sets before the black-box optimization process. If the subspace (a) achieves the highest accuracy, we consider that the criterion has successfully identified the best subspace. Otherwise, we deem that the criterion has failed to find the best subspace. For each sample size, experiments are repeated on 100 different randomly sampled development sets, and we present the success rate in Figure~\ref{fig:auto1}. The experimental results indicate that the criterion can almost always identify the best subspace if the development set contains 64 or more samples, demonstrating that the proposed criterion is effective for automatic subspace selection.

\section{Related Work}
This work is closely related to two lines of research: (a) black-box prompt tuning and (b) meta-learning for prompt tuning.

\subsubsection{Black-Box Prompt Tuning} Sun \emph{et al.}~\cite{sun2022blackbox} first introduce BBT, a method for learning continuous prompts without computing derivatives. They achieve this by sampling a random projection matrix from uniform distributions to map prompts into low-dimensional subspaces. Sun \emph{et al.}~\cite{sun2022bbtv2} improve upon this with BBTv2, which samples projection matrices from model-related normal distributions and learns deep continuous prompts. Similar to BBT, our method also optimizes prompts in low-dimensional subspaces. However, our method learns subspaces instead of randomly sampling them. Specifically, we use meta-learning on similar source tasks to learn subspace parameters, which demonstrates greater versatility across target tasks and LLMs.

\subsubsection{Meta-Learning for Prompt Tuning} The objective of meta-learning is to train a model that can adapt to new tasks with only a small number of training examples~\cite{finn2017modelagnostic}. Huang \emph{et al.}~\cite{huang2022learning} introduce MetaPT, which employs meta-learning to pre-train prompts on auxiliary tasks. Hou \emph{et al.}~\cite{hou2022metaprompting} propose MetaPrompting, which uses MAML to find a better prompt initialization. Our work is also influenced by meta-learning. However, unlike MetaPT and MetaPrompting, we present a novel meta-learning algorithm that identifies suitable subspaces for enhancing the versatility of black-box prompt tuning. With the assistance of our proposed algorithm, learning prompts without computing derivatives can achieve competitive performance.

\section{Conclusion and Future Work}

In this work, we introduce BSL, a framework that enhances the versatility of black-box prompt tuning across tasks and LLMs. To learn suitable subspaces for black-box prompt tuning, we propose a novel meta-learning algorithm for learning subspaces on similar source tasks. Extensive experiments demonstrate that BSL can enhance the performance of black-box prompt tuning on various tasks.

In the future, we aim to develop more efficient and effective derivative-free optimization algorithms for prompt tuning. We also plan to apply our proposed BSL to extra-large language models such as LLaMA-65B~\cite{touvron2023llama} to further explore the effectiveness of BSL.

\appendices

\section{Details about Update Equations}\label{sec:second_order}

In this section, let $\bm{p}_i$ denote $\bm{W}\bm{q}_i+\bm{p}_0$, and $L_1(\cdot)$ and $L_2(\cdot)$ denote $\mathcal{L}\left(\mathcal{D}_i;\cdot\right)$ and $\mathcal{L}\left(\mathcal{D}_i^{\prime};\cdot\right)$, respectively.

In Eq.\eqref{equ:phi}, since $\bm{q}$ is initialized with $\mathbf{0}$, we can represent $\bm{q}_i$ as follows:
\begin{equation}
\begin{aligned}
    \bm{q}_i&=-\alpha{\nabla}_{\bm{q}}L_1(\bm{W}\bm{q}+\bm{p}_0) \\
    &=-\alpha\left(\frac{\partial}{\partial\bm{q}}L_1\left(\bm{W}\bm{q}+\bm{p}_0\right)\right)^{\top},
\end{aligned}
\end{equation}

\noindent where
\begin{equation}
\begin{aligned}
    &\textcolor{white}{=.}\frac{\partial}{\partial\bm{q}}L_1\left(\bm{W}\bm{q}+\bm{p}_0\right) \\
    &=\frac{\partial}{\partial\bm{p}_0}L_1\left(\bm{p}_0\right)\times\frac{\partial}{\partial\bm{q}}\left(\bm{W}\bm{q}+\bm{p}_0\right) \\
    &=\frac{\partial}{\partial\bm{p}_0}L_1\left(\bm{p}_0\right)\times\bm{W}.
\end{aligned}
\end{equation}

Then $\bm{q}_i$ can be calculated as follows:
\begin{equation}
    \bm{q}_i=-\alpha\bm{W}^{\top}\left(\frac{\partial}{\partial\bm{p}_0}L_1\left(\bm{p}_0\right)\right)^{\top}.
\end{equation}

Thus, the partial derivatives $\frac{\partial}{\partial\bm{p}_0}\mathcal{L}\left(\mathcal{D}_i^{\prime};\bm{W}\bm{q}_i+\bm{p}_0\right)$ can be represented as follows:
\begin{equation}
\begin{aligned}
    &\textcolor{white}{=.}\frac{\partial}{\partial\bm{p}_0}\mathcal{L}\left(\mathcal{D}_i^{\prime};\bm{W}\bm{q}_i+\bm{p}_0\right) \\
    &=\frac{\partial}{\partial\bm{p}_0}L_2\left(\bm{W}\bm{q}_i+\bm{p}_0\right) \\
    &=\frac{\partial}{\partial\bm{p}_i}L_2\left(\bm{p}_i\right) \times \frac{\partial}{\partial\bm{p}_0}\left(\bm{W}\bm{q}_i+\bm{p}_0\right) \\
    &=\frac{\partial}{\partial\bm{p}_i}L_2\left(\bm{p}_i\right) \times \left(\bm{W}\frac{\partial\bm{q}_i}{\partial\bm{p}_0}+\bm{I}\right) \\
    &=\frac{\partial}{\partial\bm{p}_i}L_2\left(\bm{p}_i\right) \times \left(\alpha\bm{W}\bm{W}^{\top}\frac{\partial^2}{\partial\bm{p}_0^2}L_1\left(\bm{p}_0\right)+\bm{I}\right).
\end{aligned}
\end{equation}
Therefore, the calculation process of the gradient ${\nabla}_{\bm{p}_0}\mathcal{L}\left(\mathcal{D}_i^{\prime};\bm{W}\bm{q}_i+\bm{p}_0\right)$ in Eq.\eqref{equ:second_order_1} requires second-order derivatives.

To calculate the gradient ${\nabla}_{\bm{W}}\mathcal{L}\left(\mathcal{D}_i^{\prime};\bm{W}\bm{q}_i+\bm{p}_0\right)$ in Eq.\eqref{equ:second_order_2}, we represent $\bm{W}$ as $\bm{W}=(\bm{w}_1,\dots,\bm{w}_{d})$, where $\bm{w}_1,\dots,\bm{w}_{d}$ are column vectors. We also represent $\bm{q}_i$ as follows:
\begin{equation}
    \bm{q}_i=\left(q_i^{(1)},\dots,q_i^{(d)}\right)^{\top}.
\end{equation}

Thus, the gradient ${\nabla}_{\bm{W}}L\left(\mathcal{D}_i^{\prime};\bm{p}_0+\bm{W}\bm{q}_i\right)$ can be represented as follows:
\begin{equation}
\begin{aligned}
    &\textcolor{white}{=.}{\nabla}_{\bm{W}}L\left(\mathcal{D}_i^{\prime};\bm{W}\bm{q}_i+\bm{p}_0\right) \\
    &={\nabla}_{\bm{W}}L_2\left(\bm{W}\bm{q}_i+\bm{p}_0\right) \\
    &=\big({\nabla}_{\bm{w}_1}L_2\left(\bm{W}\bm{q}_i+\bm{p}_0\right), \\
    &\textcolor{white}{......}\dots, {\nabla}_{\bm{w}_d}L_2\left(\bm{W}\bm{q}_i+\bm{p}_0\right)\big) \\
    &=\Bigg(\left(\frac{\partial}{\partial \bm{w}_1}L_2\left(\bm{W}\bm{q}_i+\bm{p}_0\right)\right)^{\top}, \\
    &\textcolor{white}{......}\dots, \left(\frac{\partial}{\partial \bm{w}_d}L_2\left(\bm{W}\bm{q}_i+\bm{p}_0\right)\right)^{\top}\Bigg).
\end{aligned}
\end{equation}

For any integer $j\in[1,d]$, we calculate $\frac{\partial}{\partial \bm{w}_j}L_2\left(\bm{W}\bm{q}_i+\bm{p}_0\right)$ as follows:
\begin{equation}
\begin{aligned}
    &\textcolor{white}{=.}\frac{\partial}{\partial \bm{w}_j}L_2\left(\bm{W}\bm{q}_i+\bm{p}_0\right) \\
    &=\frac{\partial}{\partial\bm{p}_i}L_2\left(\bm{p}_i\right) \times \frac{\partial}{\partial\bm{w}_j}\left(\sum\limits_{j=1}^{d}q_i^{(j)}\bm{w}_j+\bm{p}_0\right) \\
    &=\frac{\partial}{\partial\bm{p}_i}L_2\left(\bm{p}_i\right) \times q_i^{(j)}\bm{I} \\
    &=q_i^{(j)}\frac{\partial}{\partial\bm{p}_i}L_2\left(\bm{p}_i\right).
\end{aligned}
\end{equation}
Since the calculation process of both $q_i^{(j)}$ and $\frac{\partial}{\partial\bm{p}_i}L_2\left(\bm{p}_i\right)$ only requires first-order derivatives, the calculation process of Eq.\eqref{equ:second_order_2} does not require second-order derivatives.

\bibliographystyle{IEEEtran}
\bibliography{custom}

\end{CJK}
\end{document}